\documentclass[letterpaper, 10 pt, conference]{ieeeconf}  

\IEEEoverridecommandlockouts                              

\usepackage{amsmath}
\usepackage{amssymb}
\usepackage{graphicx}
\usepackage[export]{adjustbox}
\usepackage[caption=false, font=footnotesize]{subfig}
\usepackage[noadjust]{cite} 
\usepackage{textcomp}
\usepackage{xcolor}
\usepackage{algorithm}
\usepackage{algorithmic}
\usepackage{siunitx}
\usepackage{multirow}
\usepackage{diagbox}

\usepackage[hyphens,spaces,obeyspaces]{url}
\urldef{\trossenphantomx}\url{https://web.archive.org/web/20190610001413/https://www.trossenrobotics.com/p/PhantomX-Pincher-Robot-Arm.aspx}
\urldef{\trossencad}\url{https://grabcad.com/library/interbotix-phantomx-pincher-robot-arm-kit-mark-ii-1}

\makeatletter
\let\NAT@parse\undefined
\makeatother
\usepackage[bookmarks=true]{hyperref} 

\renewcommand{\o}{\textcolor{lightgray}{0}}

\title{\LARGE \bf
A Hybrid Cable-Driven Robot for Non-Destructive Leafy Plant Monitoring and Mass Estimation using Structure from Motion
}

\author{Gerry Chen$^{1}$, Harsh Muriki$^{1}$, C\'{e}dric Pradalier$^{2}$, Yongsheng Chen$^{3}$, and Frank Dellaert$^{1}$
\thanks{This material partially supported by the
National Science Foundation (Award No. 2008302),
U.S. Department of Agriculture (Award No. 2018-68011-28371), and
National Science Foundation--U.S. Department of Agriculture Award No. 2020-67021-31526.}
\thanks{$^{1}$Institute for Robotics and Intelligent Machines, Georgia Tech, Atlanta, USA
        {\tt\small \{gchen328, vmuriki3, fd27\}@gatech.edu}}%
\thanks{$^{2}$CNRS IRL 2958, Georgia Tech Lorraine, France
        {\tt\small cedric.pradalier@georgiatech-metz.fr}}%
\thanks{$^{3}$School of Civil and Environmental Engineering, Georgia Tech, Atlanta, USA
        {\tt\small yongsheng.chen@ce.gatech.edu}}%
\vspace*{1em}  
}

\begin{document}

\maketitle
\thispagestyle{empty}
\pagestyle{empty}


\begin{abstract}
We propose a novel hybrid cable-based robot with manipulator and camera for high-accuracy, medium-throughput plant monitoring in a vertical hydroponic farm and, as an example application, demonstrate non-destructive plant mass estimation.
Plant monitoring with high temporal and spatial resolution is important to both farmers and researchers to detect anomalies and develop predictive models for plant growth.  The availability of high-quality, off-the-shelf structure-from-motion (SfM) and photogrammetry packages has enabled a vibrant community of roboticists to apply computer vision for non-destructive plant monitoring.
While existing approaches tend to focus on either high-throughput (e.g. satellite, unmanned aerial vehicle (UAV), vehicle-mounted, conveyor-belt imagery) or high-accuracy/robustness to occlusions (e.g. turn-table scanner or robot arm), we propose a middle-ground that achieves high accuracy with a medium-throughput, highly automated robot.
Our design pairs the workspace scalability of a cable-driven parallel robot (CDPR) with the dexterity of a 4 degree-of-freedom (DoF) robot arm to autonomously image many plants from a variety of viewpoints.  
We describe our robot design and demonstrate it experimentally by collecting daily photographs of 54 plants from 64 viewpoints each.
We show that our approach can produce scientifically useful measurements, operate fully autonomously after initial calibration, and produce better reconstructions and plant property estimates than those of over-canopy methods (e.g. UAV).
As example applications, we show that our system can successfully estimate plant mass with a Mean Absolute Error (MAE) of 0.586g and, when used to perform hypothesis testing on the relationship between mass and age, produces p-values comparable to ground-truth data (p=0.0020 and p=0.0016, respectively).

\end{abstract}

\section{INTRODUCTION}
Non-destructive methods for estimating plant properties are of interest to both researchers and farmers to maximize crop yields and minimize resource utilization by developing more accurate growth models and monitoring plant health more precisely \cite{Cohen22est_dynamicallyControllerEnvironmentAgriculture}.
While traditional methods of measuring plant properties require harvesting the plant (e.g. placing on a scale or performing a nutrient analysis) \cite{Atefi21fps_review_plant_phenotyping}, harvesting the entire plant makes it impossible to measure the same plant multiple times.  Instead, many ``replicates'' must be planted to (1) harvest periodically and (2) obtain a larger sample size per harvest to compensate for plant variation, both of which make developing plant models labor and resource inefficient.  Non-destructive methods, which are defined by not requiring harvesting, for estimating plant mass (and other properties) with both high throughput and accuracy would enable more accurate plant modelling \cite{Rahaman15fps_phenotyping_for_plant_growth,Tariq20chapter_phenotyping}.

Significant work has applied computer vision and/or robotics to non-destructive plant analysis and can be broadly categorized into 2D and 3D approaches \cite{Li14sensors_review_imaging_techniques}.
Works focusing on aggregate field biomass rather than individual plant mass \cite{Xiaowei15rs_satellite_compare_different_sources} do not achieve comparable accuracies to those dedicated to individual plants, so we focus on the latter.
2D computer vision approaches with single-plant resolution monitor plant disease, health, and canopy area well \cite{Li11rs_satellite_LAI_riceyield}, but struggle to estimate mass with high accuracy \cite{Aich:2018wacv}: a critical component of growth modeling.
3D approaches include using RGB \cite{Carlone15icra_towards_4Drecon}, Stereo \cite{Ni:2016}, Time-of-Flight-based \cite{Hosoi06tgrs_lidar_voxel_individual_trees,Tilly14ars_laser_scanning}, and a number of other specialized sensors to estimate plant properties such as mass, dimensions \cite{Blok21be_rgb_stereo_brocolli}, and organ structure \cite{Elias22icra_uav_plants_templatematching,Shi19be_plant_segmentation_3d}.
In most cases, multiple viewpoints are required to accurately assess the 3D structure of a plant, with occlusions posing particular difficulty \cite{Blok21be_rgb_stereo_brocolli,Elias22icra_uav_plants_templatematching}.  Existing approaches tend to focus on either high-throughput or high-accuracy.
High-throughput approaches, such as satellite \cite{Bannari95rs_remoteSensing_VegetationIndices,Mulla13be_remoteSensingSurvey}, UAV \cite{Dandois10rs_uav_rgb,Geipel14rs_uav_cropSurfaceModel,Bendig15ijaeog_uav_cropSurfaceModel, Elias22icra_uav_plants_templatematching}, conveyor \cite{Li20fbb_high_throughput_review}, and cart \cite{Xu22pp_plant_ground_robot_phenotyping_review,Lumme08_cart_lidar,Van-Der-Heijden:2012wv,Polder:2014,Dong17icra_4d_tractor} -based systems, can phenotype large numbers of plants but are less capable of handling occlusions due to their limited viewing angles.
Meanwhile, high-accuracy approaches collect data from many viewpoints using e.g. robot arms/gantries \cite{Wu19ral_robot_arm_phenotyping,Lee18pone_gantry_segmentation}, scanner turntables \cite{Demby21ssci_4d_stereo}, and handheld scanners/photogrammetry \cite{Hosoi06tgrs_lidar_voxel_individual_trees,Ni:2016}, but are not designed to image large numbers of plants in an automated fashion.

\begin{figure}
  \centering
  \includegraphics[height=0.4428834835\linewidth]{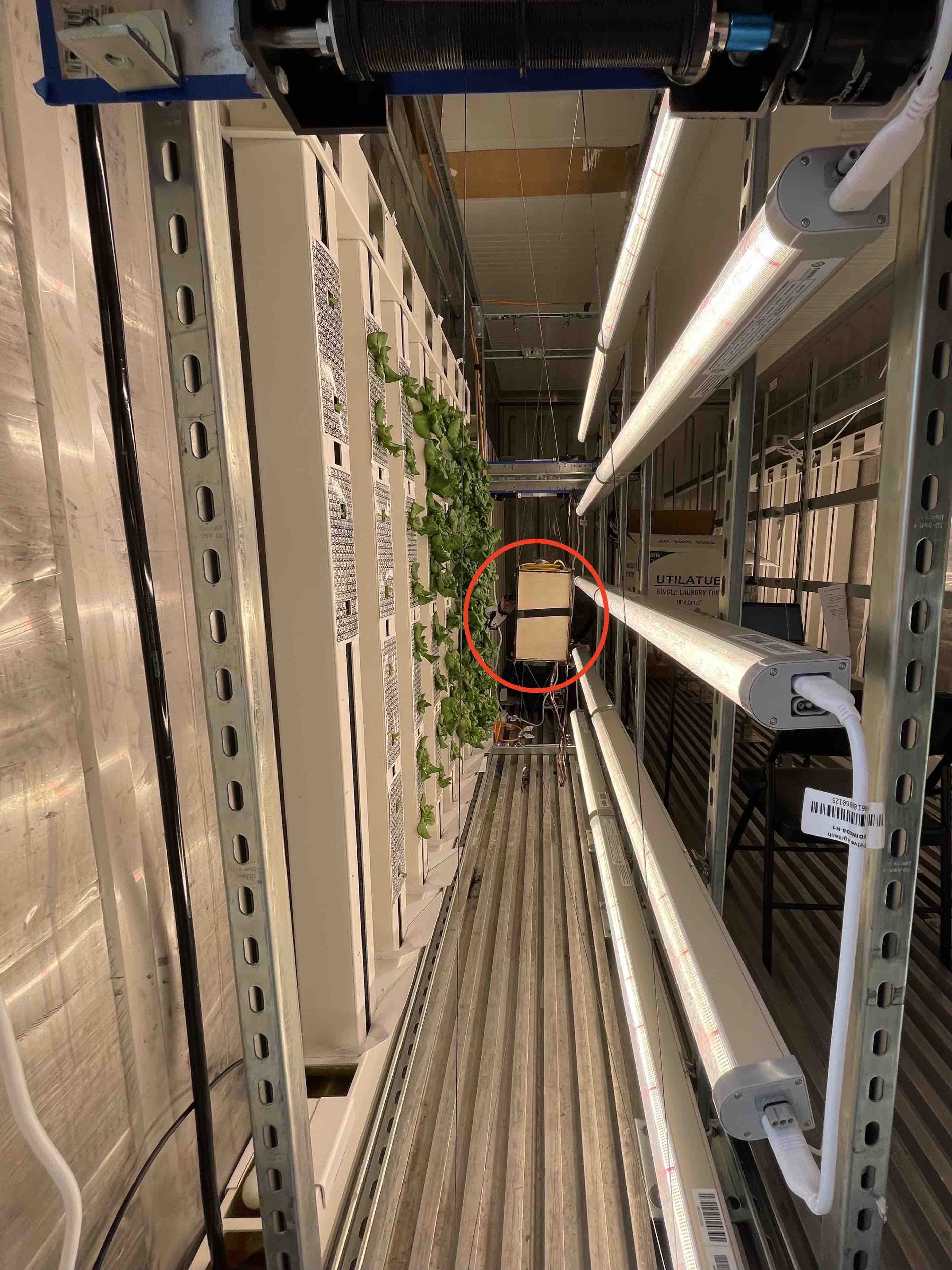}
  \includegraphics[height=0.4428834835\linewidth]{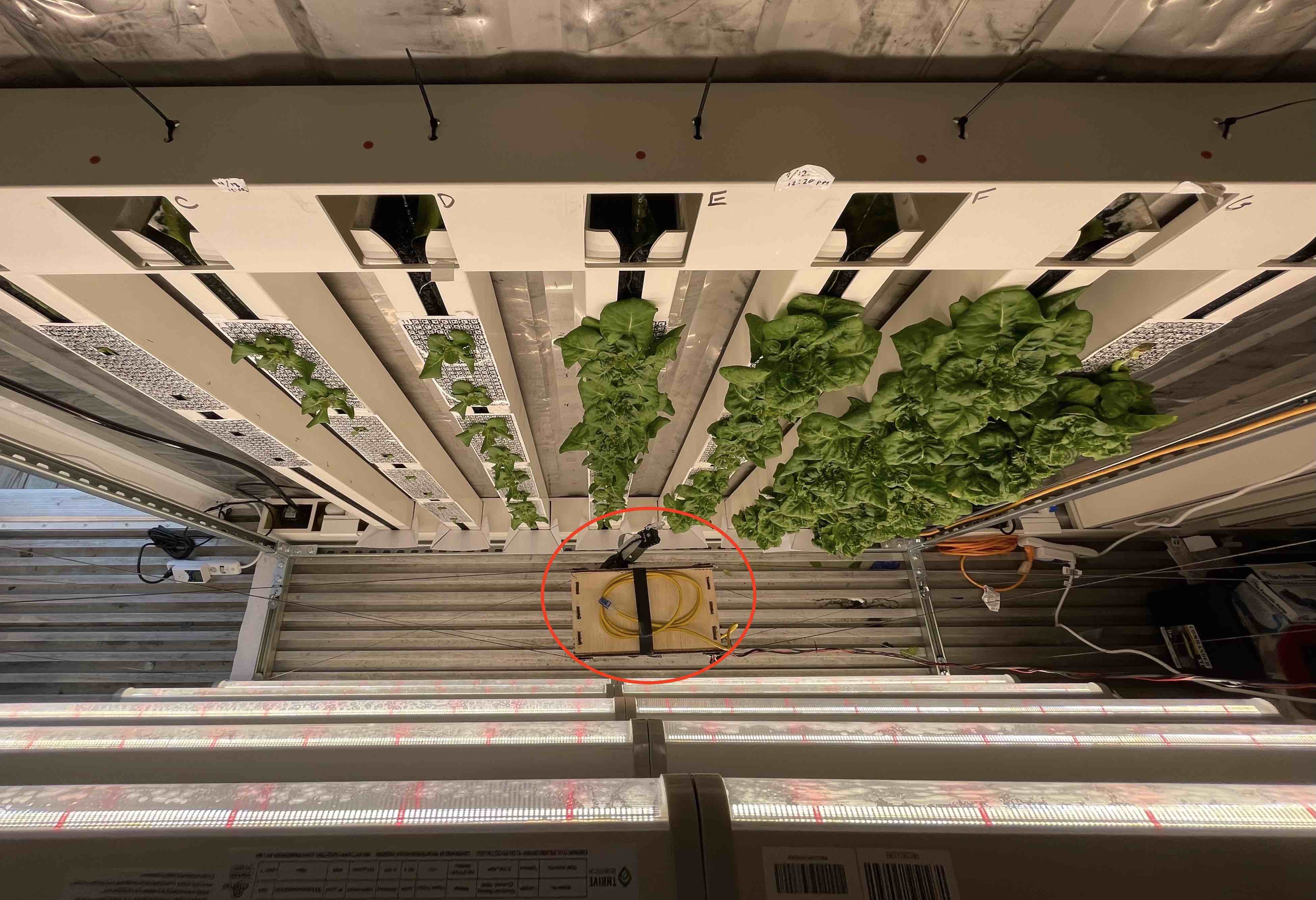}
  \caption{The robot is placed in a vertical indoor grow chamber between the plants and the grow lights (Left: side view; Right: top view).  The robot consists of a CDPR which moves the robot arm (circled in red) from plant to plant to take photos from a wide set of viewpoints.}
  \label{fig:system_with_plants}
\end{figure}

In this work, we propose a medium-throughput, high-accuracy approach that pairs the large workspace of a cable robot with the dexterity of a robot arm manipulator to collect photos of many plants from many viewpoints.
We validated the hybrid cable-manipulator approach with a 3\SI{}{\m}\,$\times$\,2\SI{}{\m} cable-driven parallel robot (CDPR) and 4 degree-of-freedom (DoF) manipulator moving an RGB camera to image lettuce plants in the vertical hydroponic system shown in Fig. \ref{fig:system_with_plants}.  We collected a dataset consisting of photos, fresh masses, and dehydrated masses of 54 plants.  Finally, we demonstrated the scientific utility of the data collected by our robot both qualitatively and quantitatively by comparing dense 3D reconstructions, mass estimates, hypothesis tests, and occlusions against 3 baselines based on simulated photos for competing approaches.

\section{APPROACH}

The robot used for collecting photographic data of the lettuce plants consists of 2 subsystems: a robot arm mounted on the end effector of a cable-driven parallel robot (CDPR).
The purpose of the robot arm is to collect large numbers of photos of a plant from various, repeatable angles for use in SfM and other analysis techniques.  Meanwhile, the CDPR enables analyzing a larger quantity of plants by moving the robot arm from plant to plant, enlarging the workspace of the robot arm to cover dozens of plants.

\subsection{Mechanical Design}
\subsubsection{CDPR}
The cable-based robot platform is chosen for its scalability and economy \cite{Kirchgessner17fpb_cablerobot_phenotyping,Bai19cea_cablerobot_agriculture}, which allow the robot to reach multiple plants and remain permanently installed for complete autonomy.
Although our demonstrated robot is only 2.9m\,$\times$\,2.3m in size, in principle it can scale to almost any size vertical grow towers.  As compared to e.g. gantry or conveyor type systems, cable-based systems remain almost constant in price relative to size.  The planar design is chosen for its favorable tradeoff between capability and cost/complexity, since collisions with plants would limit the utility of out-of-plane motions anyway.

\begin{figure}
  \centering
  \includegraphics[height=0.4239417642\linewidth]{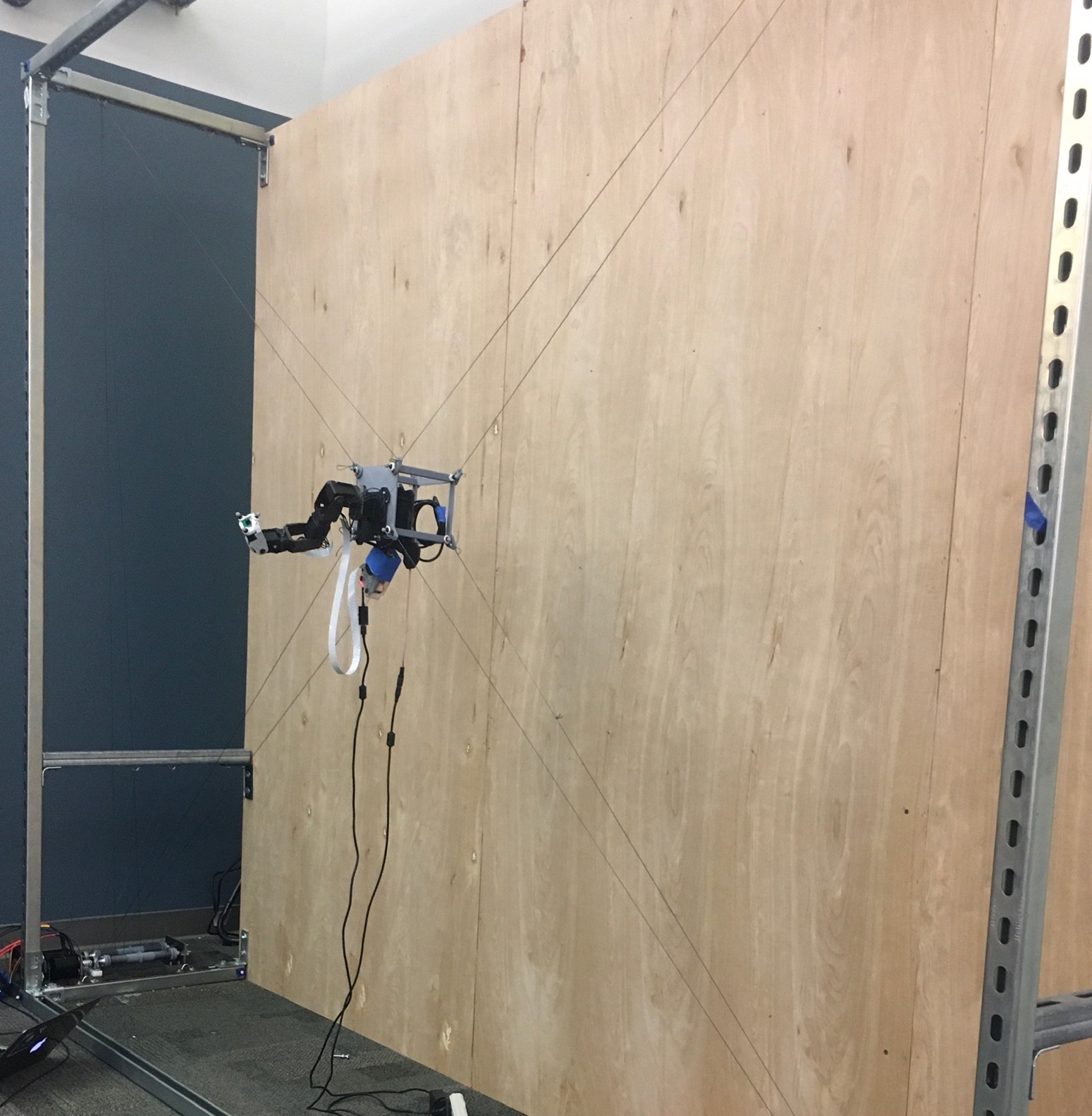}
  \includegraphics[height=0.4239417642\linewidth]{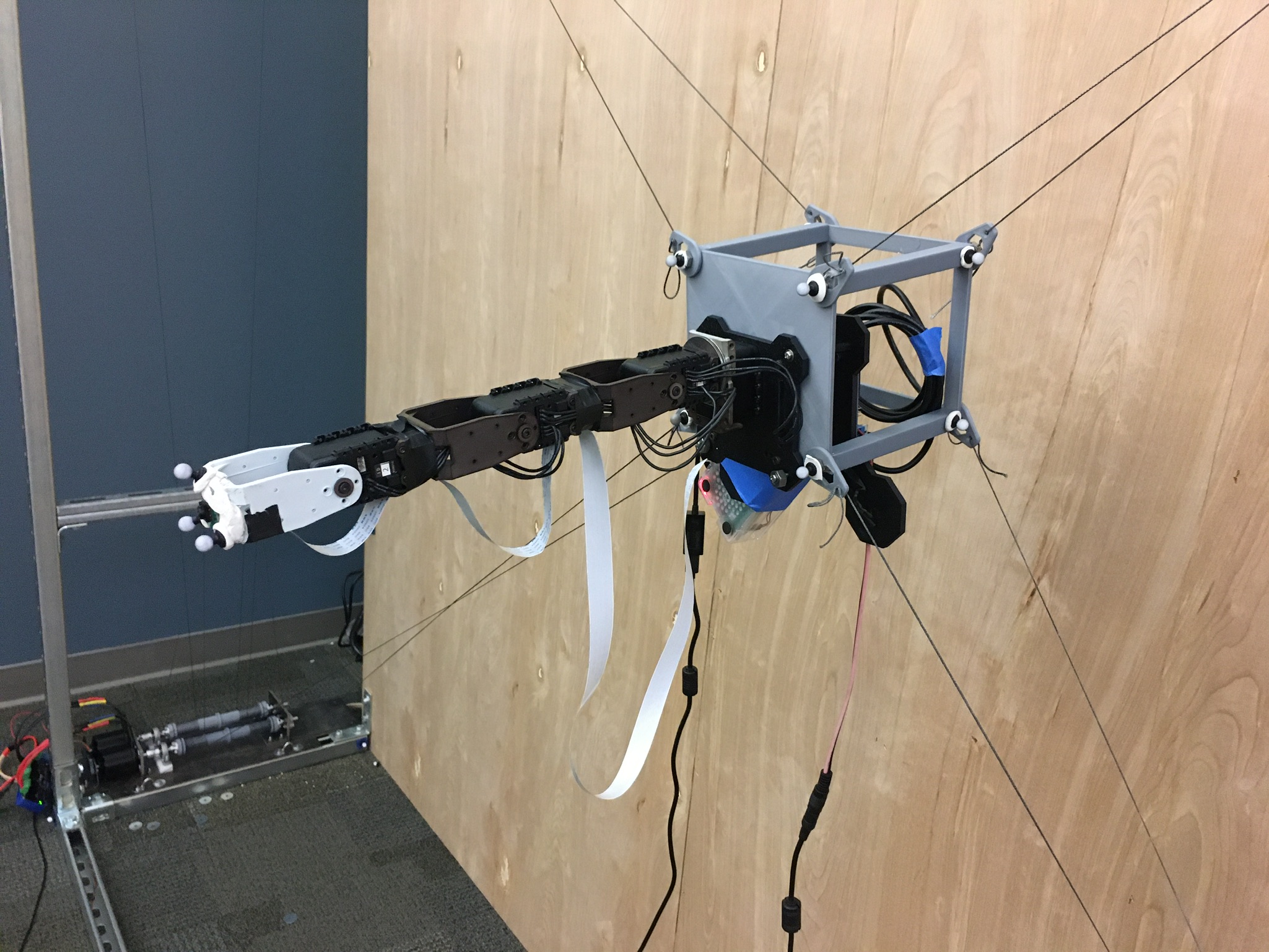}
  \caption{Left: CDPR consists of 4 pairs of cables controlling a moving platform on which a robot arm is mounted.  Right: Each pair of cables is crossed provide additional out-of-plane stability, but both cables in each pair are driven by the same motor.}
  \label{fig:doubled_cable} \label{fig:cdpr}
\end{figure}

\begin{figure}
  \centering
  \includegraphics[width=0.49\linewidth,valign=t]{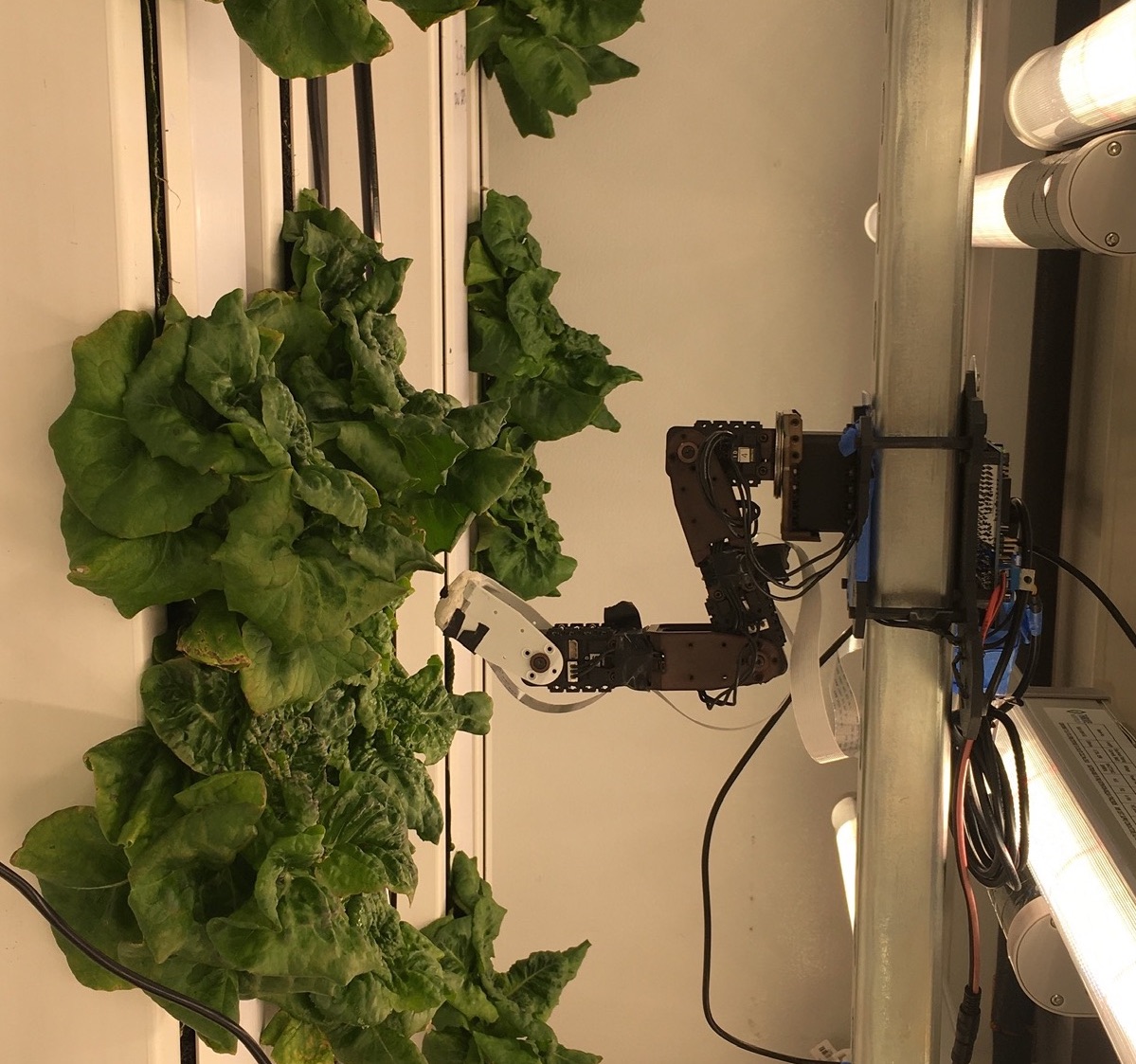}
  \includegraphics[width=0.49\linewidth,valign=t]{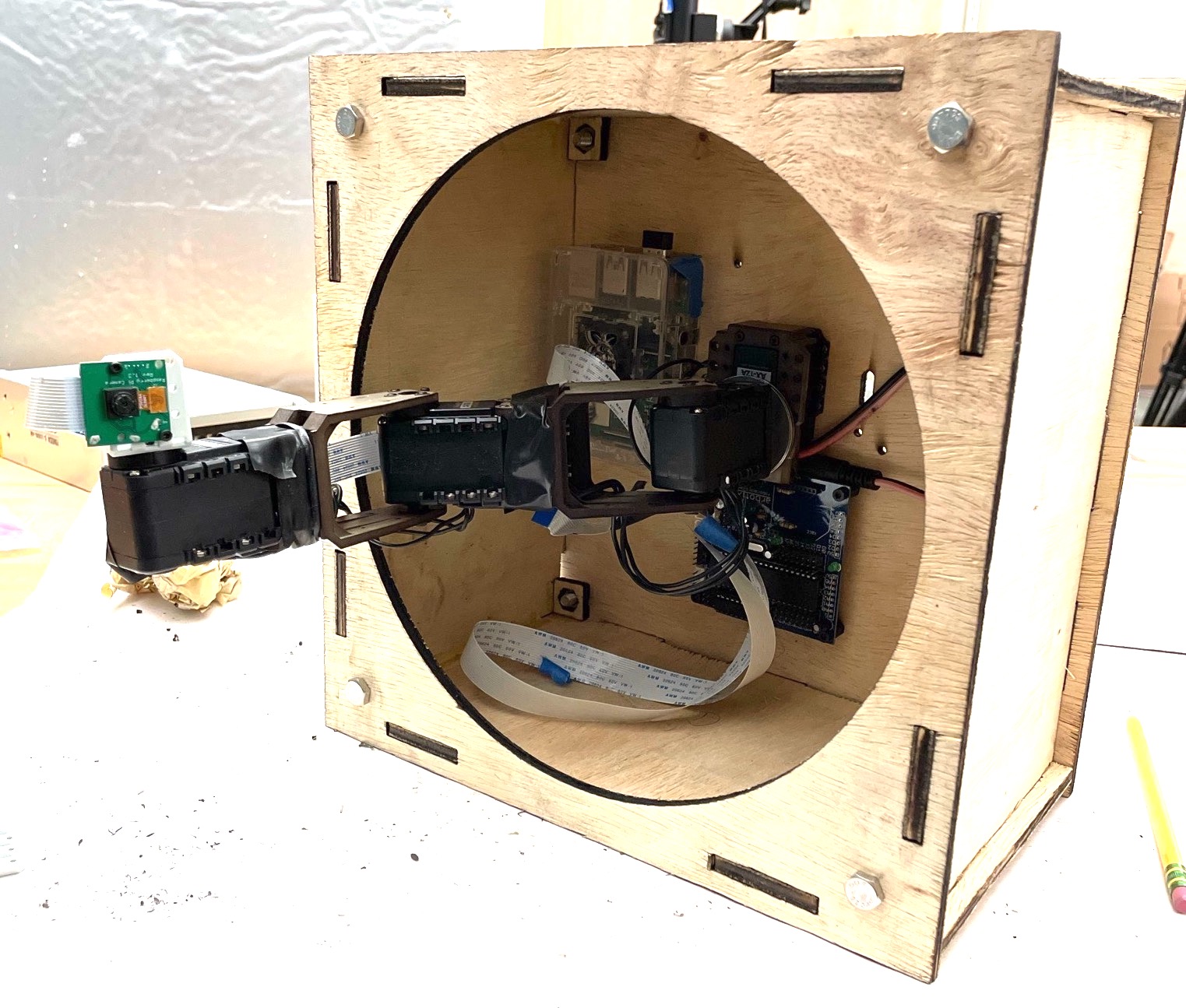}
  \caption{4DoF Robot arm with camera used to take a large number of photos from various angles of a single plant.  Left: Arm (without wooden cover or CDPR) taking photos of a plant.  Right: Arm inside wooden protective cover.}
  \label{fig:arm_photo}
\end{figure}

The CDPR is an 8-cable, 4-motor planar CDPR with a workspace of roughly 2.9m\,$\times$\,2.3m.  
Details on the design can be found in \cite{Chen22icra_GTGraffiti}, with the primary distinctions being that (a) the robot arm shown in Fig. \ref{fig:arm_photo} is used in place of the spray paint carriage, and (b) the cables are doubled to provide more out-of-plane stability.  The doubled cables consist of two cables spooled with two drums on a shared shaft driven by a single motor, as depicted in Fig. \ref{fig:doubled_cable}.
Although certain 4-motor planar CDPR geometries can control both translation \emph{and rotation} in the plane, we choose a geometry which largely precludes rotational motion.  This choice was made because the benefit of the additional stiffness enabled by the chosen geometry outweighs the lack of CDPR rotation, especially when coupled with the robot arm's shoulder joint.
The CDPR with robot arm is shown in Fig. \ref{fig:cdpr}.


\subsubsection{Robot Arm}
The robot arm (Fig. \ref{fig:arm_photo}) is chosen to supplement the planar CDPR with the dexterity to reach around a plant and take photos from a variety of viewpoints.
The robot arm was chosen to have 4DoF in a configuration that, combined with the 2DoF of the CDPR, provides the robot with full SE(3) motion.
Although in theory 1DoF is redundant with camera-axis rotation and another 2DoF are unnecessary with a sufficiently wide field-of-view camera, we found in practice that they are helpful when reaching around plants to avoid collisions with neighboring plants.

\begin{figure}
  \centering
  \subfloat[Top view\label{fig:coords_top}]{%
  \includegraphics[scale=0.18, page=1,trim=0 3.39in 6.99in 0, clip, valign=c]{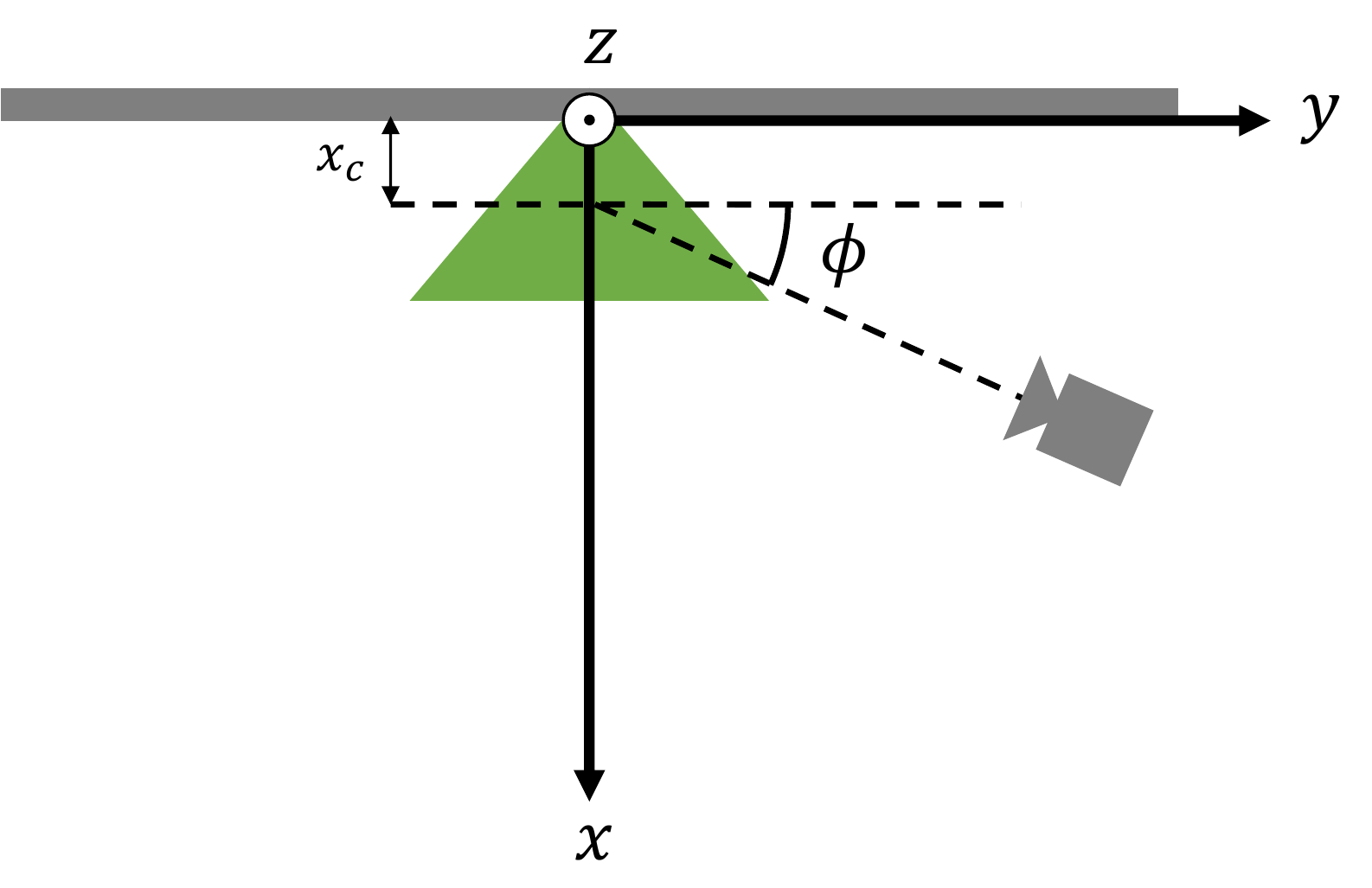}
    \vphantom{
    \includegraphics[scale=0.18, page=2,trim=0 0in 8.59in 0, clip, valign=c]{figs/figs.pdf}}
  } \hspace*{4em}
  \subfloat[Front view\label{fig:coords_front}]{%
  \includegraphics[scale=0.18, page=2,trim=0 0in 8.59in 0, clip, valign=c]{figs/figs.pdf}
  }
  \caption{Coordinate frame of the camera with respect to a lettuce plant.}
  \label{fig:coords}
\end{figure}

The robot arm is adapted from a Trossen Robotics PhantomX Pincher Mark II \cite{trossen_arm}, which is a 4DoF robot manipulator using Dynamixel AX-12A servos.  The links were extended to have lengths of \SI{0.107}{\m}, \SI{0.194}{\m}, and \SI{0.032}{\m} after the shoulder, elbow, and wrist joints respectively.
We also replaced the gripper with a Raspberry Pi Camera Module v2, which uses a IMX219 8MP sensor.  The 4 DoF allow rotation in $\theta$ with the base joint and both translation and rotation in the $x-r$ plane (see Fig. \ref{fig:coords}).  The completed robot arm is shown in Fig. \ref{fig:arm_photo}.

\subsection{Electrical and Communication Design}

\begin{figure}
  \centering
  \includegraphics[width=\linewidth, page=3,trim=0 0.72in 2.49in 0,clip]{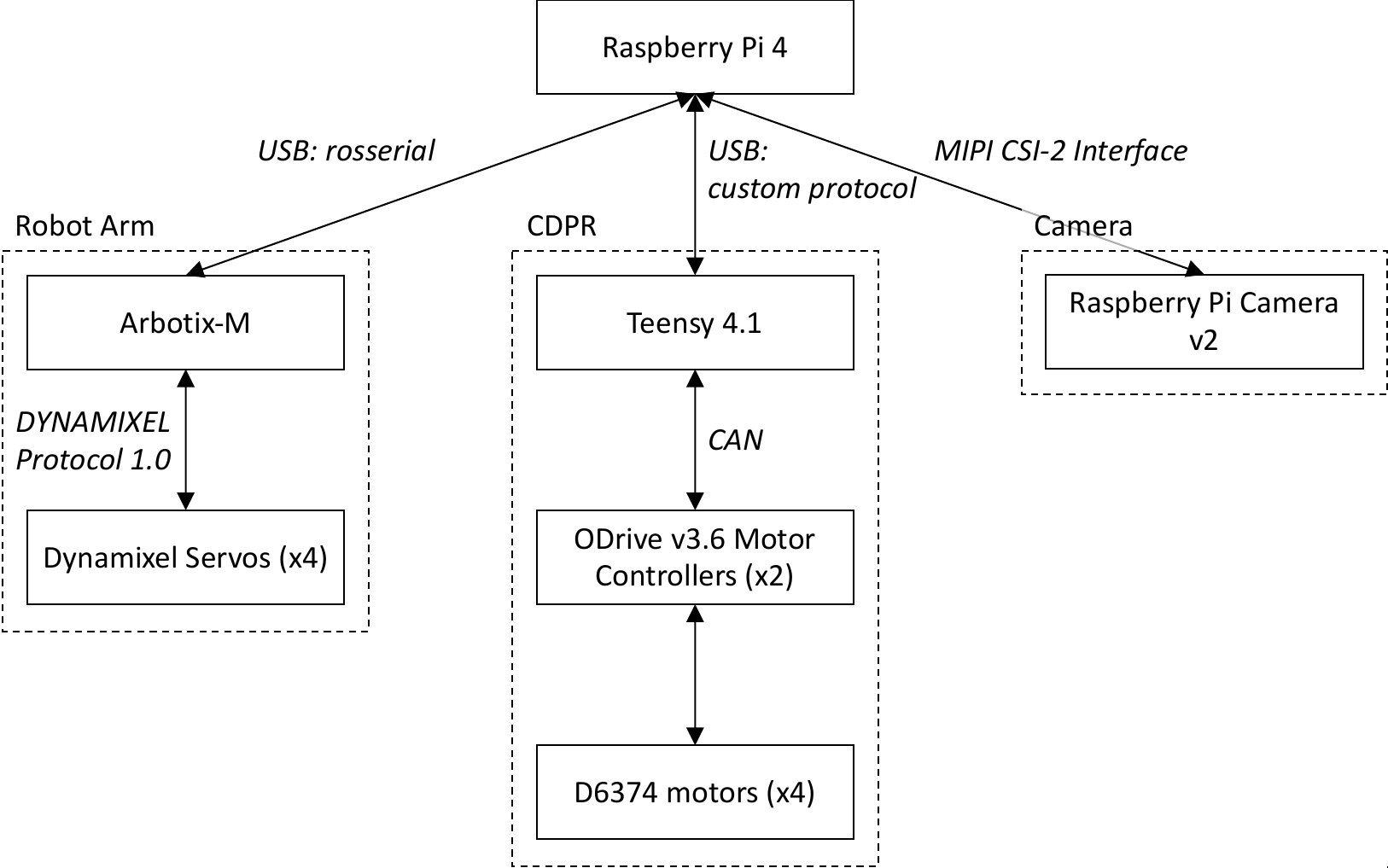}
  \caption{System communication overview.}
  \label{fig:communication}
\end{figure}

\begin{figure}
  \centering
  \includegraphics[width=0.7\linewidth]{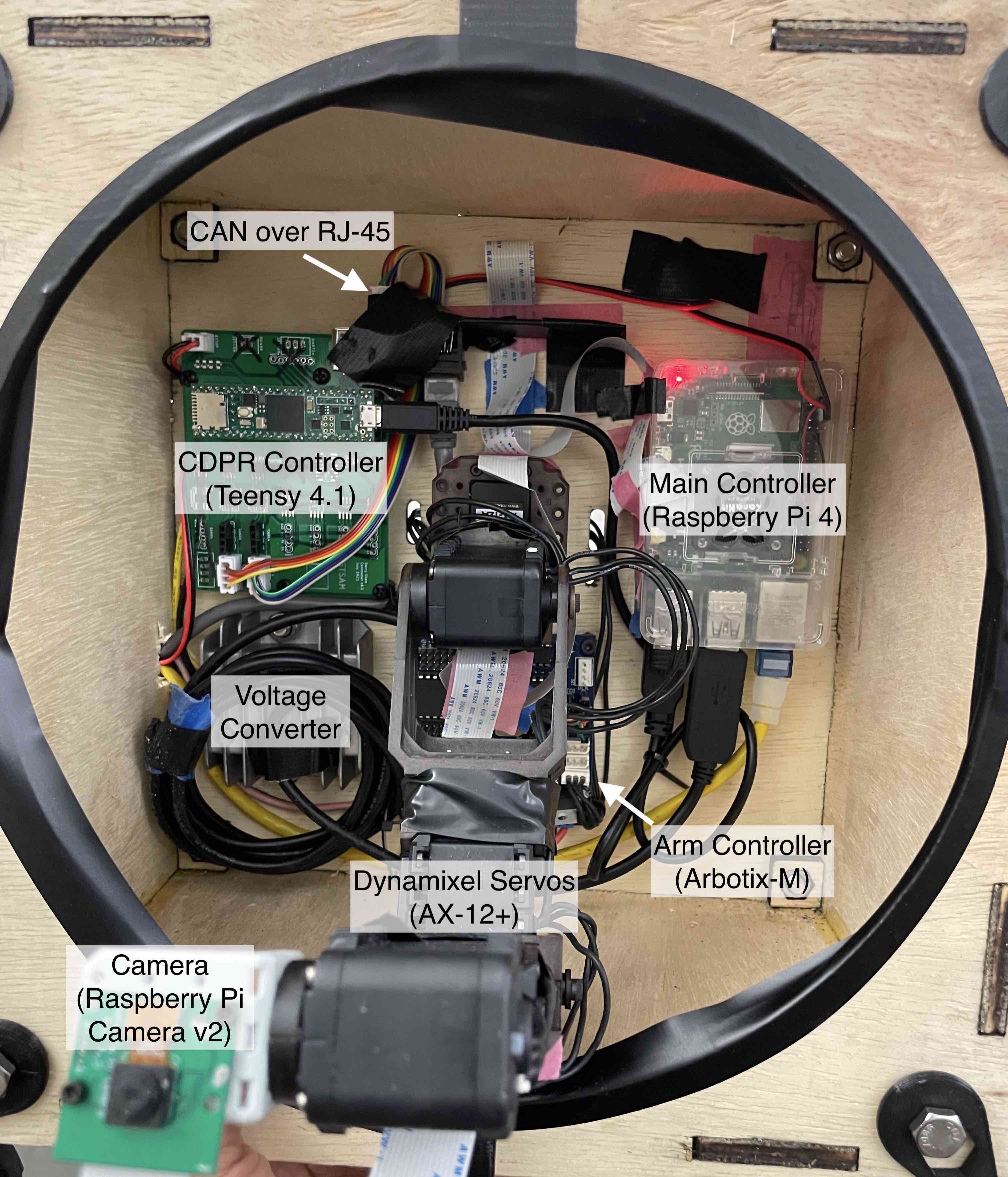}
  \caption{Electronics mounted on the CDPR moving platform inside a protective wooden cover.}
  \label{fig:arm_box_annotated}
\end{figure}

A Raspberry Pi 4 controls the camera, robot arm, and CDPR using ROS, as overviewed in Fig. \ref{fig:communication}.  The electronics are shown in Fig. \ref{fig:arm_box_annotated}, excepting the motor controllers and motors which are available in \cite[Fig. 6 (right)]{Chen22icra_GTGraffiti}.
The camera is connected and controlled directly by the Pi.
The arm is controlled by an Arbotix-M microcontroller which receives joint angle position commands from the Pi and sends back joint angle position feedback.
The CDPR is controlled by a Teensy 4.1 which receives high-level cartesian position commands from the Pi and applies low-level motor torque commands to the motor controllers using the algorithm from \cite[Sec. III.A]{Chen22iros_lqg_cdpr} using factor graphs \cite{Dellaert17fnt_factorgraphsforrobotperception,Chen19blog_lqr-blogpost,Yang21icra_ecLQR}.

\subsection{Data Collection Algorithm}

The algorithm used for a data collection session consists of using the CDPR to move to a plant, then using the robot arm to take photos from a variety of viewpoints.
The positions of the plants on the grow towers are known and pre-programmed for the CDPR to move to, and the set of camera poses is adjusted based on the age of the plant: the view angles remain consistent while the distance to the plant increases with plant age.

\section{EXPERIMENTAL METHODS}
We validate our robot design and data collection procedure by collecting data across 2 growth cycles with different nutrient (fertilizer) schedules, analyzing the mass estimates using SfM, and comparing against 3 baseline methods simulating high-throughput or high-accuracy approaches.
Although more modern techniques for estimating plant properties using computer vision are available, we use a simple, off-the-shelf SfM approach as it is sufficient to validate the data collection capabilities of our robot.

\subsection{Lettuce Growing and Dataset Collection Procedure} \label{ssec:grow_procedure}

The data collection procedure was designed to collect photos, masses, and elemental nutrient contents of 72 total plants distributed across various growth stages and 2 different nutrient schedules (``Experiment 1'' and ``Experiment 2'').  We used Bibb Butterhead Lettuce plants because they are economically well-suited to indoor farming, grow fast, and have qualities that make them challenging for computer vision approaches (e.g. high degree of self-occlusion and self-similar texture).

Due to learned lessons from Experiment 1, the experiments used slightly different procedures, but the growing procedure for each individual plant remained the same:
\begin{enumerate}
  \item To germinate, place seeds in a dampened rockwool substrate, and place the substrate in an incubator next to grow lights for 14 days.
  \item Transplant the seedlings (with substrate) into the vertical hydroponic grow towers (``Day 0'').
  \item Take 64 photos (1 top-down, and 21 from each of 3 rings with constant $\phi$) per day.
  \item Harvest and measure ground truth (GT) data at the scheduled time.
\end{enumerate}
The harvest process consists of cutting the plant at the base, weighing immediately (``Fresh Mass''), dehydrating for 48 hours, weighing again (``Dry Mass''), and performing a nutrient analysis.  The fresh mass must be weighed immediately since, once cut, transpiration causes the plant to lose mass so quickly that the reading on a gram scale will observably decrease during the few seconds it is being weighed.

For Experiment 1, the \textit{General Hydroponics Flora Series} fertilizer is used with ratios 3:2:1 of \textit{FloraGro}, \textit{FloraMicro}, and \textit{FloraBloom} totalling 138ml of fertilizer per 100L of water.  The pH is monitored and buffered daily, and the hydroponic system is flushed/replenished every 2 weeks.
For Experiment 2, the Modified Sonneveld's solution from \cite{Mattson14ig_hydroponic_nutrient_formula} is used.  The pH is monitored and buffered daily, and the hydroponic system is flushed/replenished 3 times per week.

For Experiment 1, 48 plants were planted in sets of 12 each week and harvested at the same time when the oldest plant reached maturity (28 days after transplant).
For Experiment 2, 24 plants were planted at the same time and harvested 3/day from 21 to 28 days after transplant.  The Experiment 2 plants were planted at the same time to reduce variability due to germination conditions.  Consequently, they reached maturity at similar times so they were planted at half the density as in Experiment 1 to reduce overcrowding.  The harvest schedule was designed to capture the most ``interesting'' portion of the logistic growth curve.

The specifications for the indoor vertical hydroponic setup will be available in \cite{Sharkey23_pilot_site_tbd}, and the vertical grow rig with cable robot is shown in Fig. \ref{fig:system_with_plants}.

\subsection{Mass Estimation using SfM} \label{ssec:mass_estimation}
We validated the efficacy of our robot and data collection by estimating the masses of the lettuces using SfM.  For each plant, we first used COLMAP to generate a dense point cloud of the plant from the photos taken by the robot.  Using robot forward kinematics as camera pose priors, we transformed the point cloud into a canonical frame to resolve monocular ambiguity.
We then applied a number of programmatic cleaning steps to discard outlier and background points from the point-cloud.  Next, we generated a mesh of the plant using Poisson Surface Reconstruction and applied Poisson Disk Sampling to make a more uniform mesh while maintaining a point density sufficient to voxelize without bias on a 3mm grid size.  We computed both (1) the surface area of the mesh and (2) a volume estimate by voxelizing the mesh at a 3mm grid size and counting the total occupied voxels.  Finally, we applied linear regressions to estimate mass using either surface area or volume.

\subsection{Baseline Methods} \label{ssec:baseline_methods}
We evaluated our approach against (a) high-throughput methods -satellite, UAV, or conveyor-belt imagery- by using subsets of the total photos collected, and (b) high-accuracy methods by using only the robot arm without the CDPR to replicate similar high-accuracy methods in the literature.

\subsubsection{Baseline 1}
Here we used only a single top-down photo of each plant.
We applied a 2-layer CNN to segment the plant (foreground) from the background in the undistorted photos and count the number of pixels occupied by the plant.
We then used the known pose of the camera relative to the base of the plant, combined with the calibrated camera intrinsics, to approximate the projected area of the plant.

\subsubsection{Baseline 2}
Here we simulated over-canopy approaches (such as UAV or conveyor-belt) by using only the photos with camera poses that do not ``reach around'' the plant.
Specifically, we set the minimum threshold for the x-position of the camera to be 17cm (the maximum observed x-dimension of any plant in the datasets) and used only photos from the dataset with camera poses beyond the threshold.  We applied the same mass estimation algorithm as in \ref{ssec:mass_estimation}.

\subsubsection{Baseline 3}
Here we replicated a low-throughput, high-accuracy method by using only the robot arm without the CDPR, instead manually placing the base of the robot arm in front of each plant.
For each plant, we commanded the robot arm to a ``top-down photo'' pose for reference, fixed the arm's base in place on a stand (see Fig. \ref{fig:arm_photo}, left), and took the same set of photos per plant we used for our method.
We compared the throughput and camera pose consistency.







\section{EXPERIMENTAL RESULTS}
Using our robot system, we produce a dataset of 54 plants, each containing 64 photos/day, fresh mass, dry mass, and elemental nutrient content.  
Of the 72 plants initially planned, 18 were discarded due to overcrowding or death.
The full datasets are available online \cite{dataset_links}.

We applied a number of quantitative and qualitative metrics on the datasets to evaluate our design:
\begin{enumerate}
  \item characterize the speed (throughput) of data collection,
  \item evaluate regressions between GT and estimated metrics (using $R^2$ and cross-validation MAE),
  \item compare statistical power using GT vs estimated metrics for hypothesis testing,
  \item estimate point cloud occlusion using a non-linear regression, and
  \item visualize point clouds.
\end{enumerate}

\subsection{Data Collection Throughput}
Our robot system was capable of autonomously collecting data at approximately 2640 photos/hour and spanning 56 plants at a density of 350 cm$^2$/plant (although the experiments had fewer viable plants than the robot was capable of imaging).  Given the inherent scalability of cable robots, increasing the size of the cable robot to reach a greater number of plants should be possible.  Additionally, higher quality cameras can dramatically increase the photo capture rate by enabling faster shutter speeds or even continuous robot arm motion (currently, the arm must stop for each photo to eliminate motion blur and rolling shutter effects; motion input-shaping may also improve the capture rate).

Whereas Baseline 3 required a pair of skilled humans to collect data and was only capable of collecting 300-600 photos/hour (64 photos/plant) (depending on the skills of the operators), during experiment 2, our robot system was demonstrated to run autonomously without human supervision across several days. We anticipate that the robot can be run 100\% autonomously in future growth cycles.

\subsection{Mass Estimation Regression Results} \label{ssec:regression}
\begin{figure}
  \centering
  \includegraphics[width=0.85\linewidth]{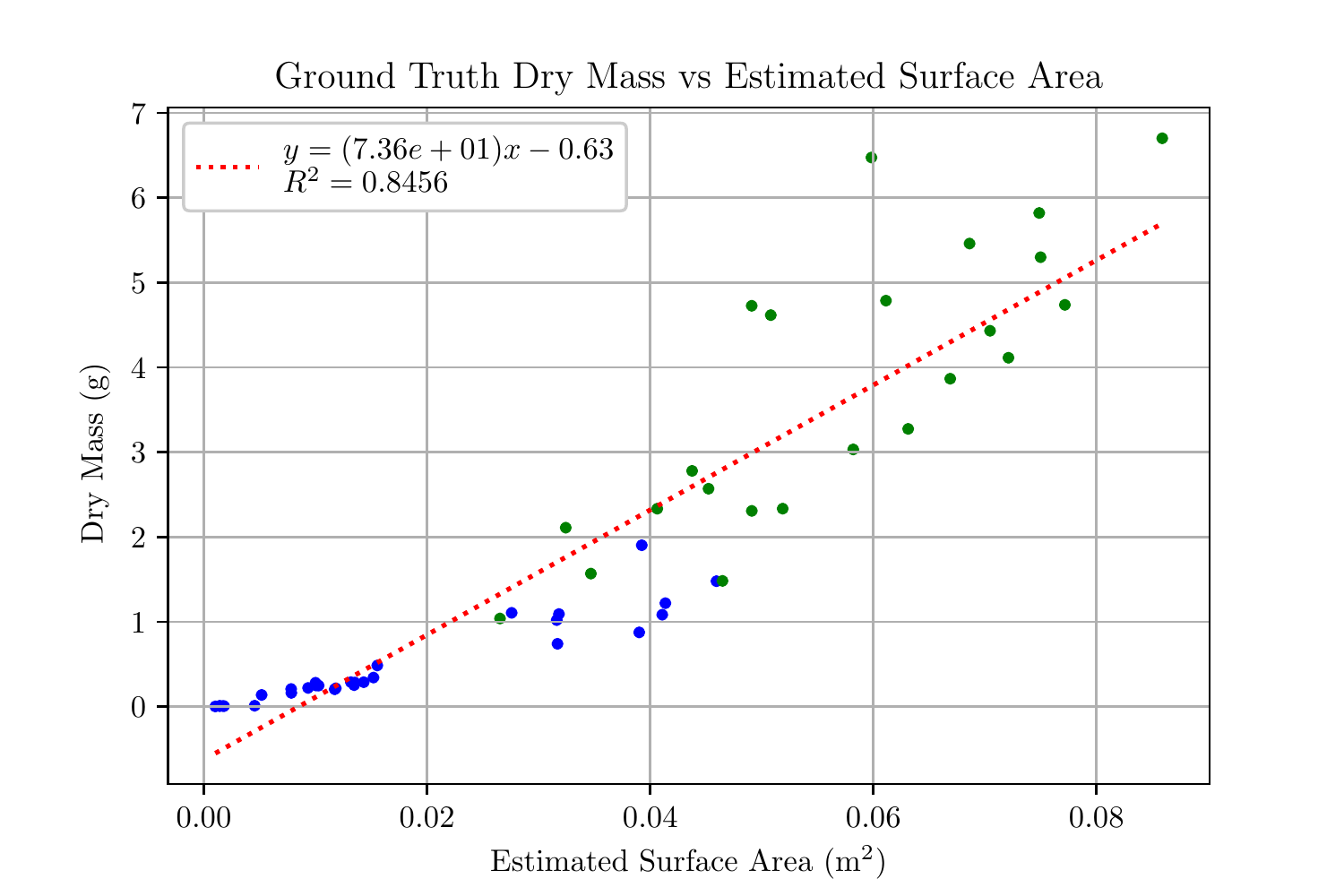}
  \includegraphics[width=0.85\linewidth]{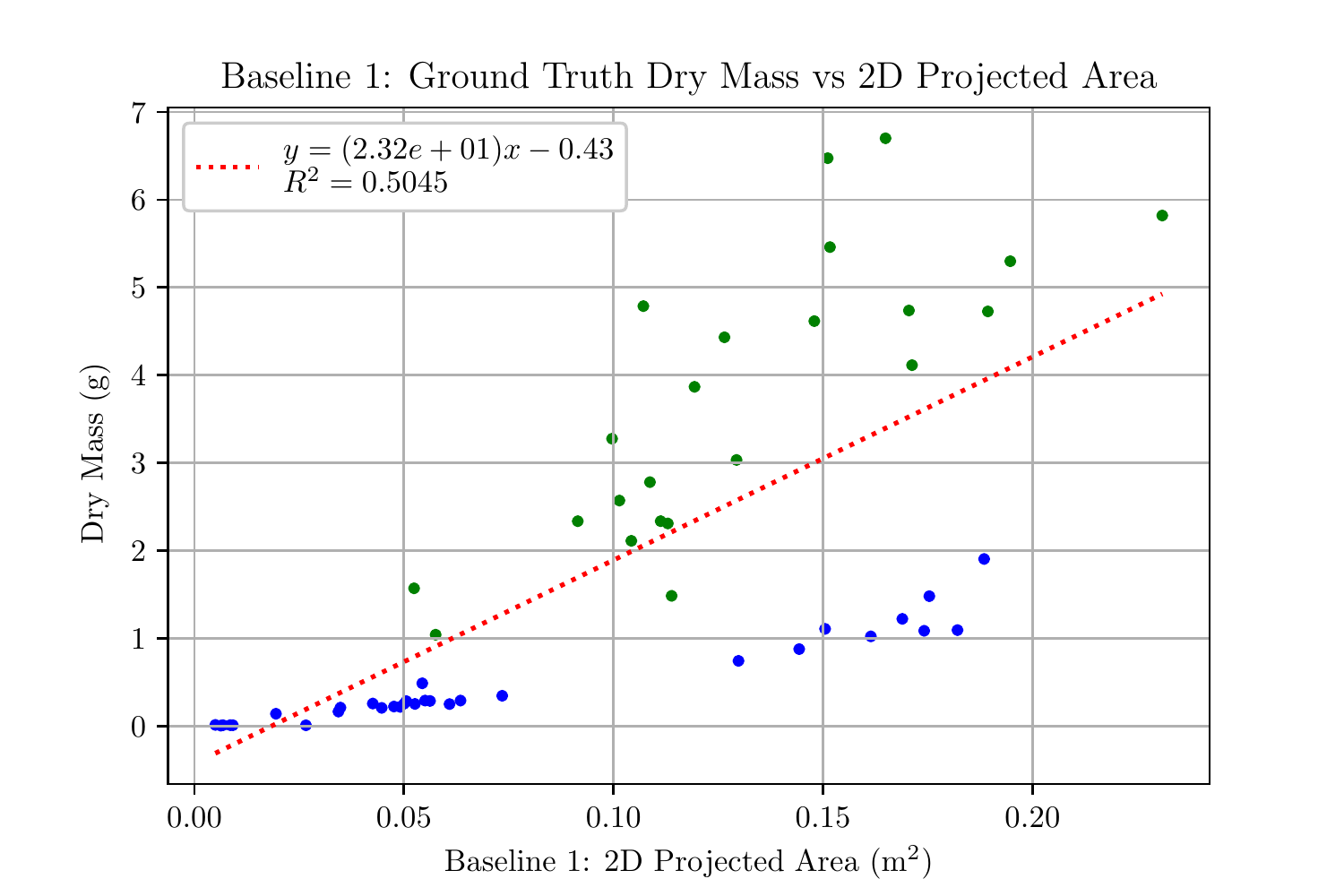}
  \caption{Sample linear regression results estimating dry mass (GT) using the surface area computed from SfM (top) and the projected area estimate from a single top-down photo of each plant (bottom).  Our regression is better than the baseline's which demonstrates that the our robot design's ability to capture many viewpoints obtains better performance than a simulated high-throughput approach with limited viewpoints of a plant.}
  \label{fig:regressions}
\end{figure}

Based on $R^2$ and leave-one-out cross-validation MAE values, our estimated metrics using the full dataset are better correlated to the GT masses than the baseline metrics.  Figure \ref{fig:regressions} shows the linear regressions and $R^2$ values for a couple representative pairs, and Table \ref{tab:regressions} shows the $R^2$ and MAE values for linear regressions between estimated metrics and ground truth masses.

\begin{table}
  \centering
  \caption{Linear Regression Results}
  \label{tab:regressions}
  \begin{tabular}{l|cc|cc}
   \multirow{2}{*}{Estimation Metric} 
                                      & \multicolumn{2}{c|}{GT: Fresh Mass} & \multicolumn{2}{c}{GT: Dry Mass} \\
    & $R^2$ & MAE (g) & $R^2$ & MAE (g) \\
    \hline
	Surface Area (ours)          & \bf0.845 & \bf11.216 & \bf0.846 & \bf0.586 \\
	Volume (ours)                & 0.833 & 11.671 & 0.832 & 0.617 \\
	Baseline 1: Projected Area   & 0.537 & 19.976 & 0.505 & 1.084 \\
	Baseline 2: Surface Area     & 0.292 & 26.049 & 0.285 & 1.401 \\
	Baseline 2: Volume           & 0.277 & 26.439 & 0.269 & 1.422 \\
  \end{tabular}
\end{table}

\subsection{Statistical Power}
Ultimately, one primary application of non-destructive phenotyping is to draw scientific conclusions from the computed metrics.  To this end, we evaluated our method by running statistical significance tests using (1) ground-truth masses, (2) estimated masses our method in \ref{ssec:mass_estimation}, and (3) estimated masses using the baseline methods from \ref{ssec:baseline_methods}.  We used ANOVA \cite{Mishra19ca_statistical_tests} to test 2 hypotheses: (a) the age of the plant is correlated with the mass of the plant and (b) nutrient schedule is correlated with the mass of the plant.

To test the hypothesis that age is correlated with mass, we grouped the plants into their harvest days and run one-way ANOVA tests to determine if the different groups have different means of each of the mass metrics.
Due to the non-uniform frequencies of ages in the harvested plants, shown in Table \ref{tab:plant_age}, we performed separate tests for Experiments 1 and 2, and for Experiment 1 we only used the 15 and 21 day-old age groups (note that 2-group ANOVA is equivalent to t-test with independent samples \cite{Mishra19ca_statistical_tests}).
The results are presented in columns 2 and 3 of Table \ref{tab:anova} and show that both our estimate volume and surface area have similar statistical powers as the GT masses ($p<0.005$), while the baselines are nearly an order of magnitude less powerful.  The exception is Baseline 1 for Experiment 1, which is likely an artifact of the camera position algorithm used for experiment 1 using different camera positions for the different age groups.

To test the hypothesis that nutrient schedule is correlated with plant mass, we tested whether Experiments 1 and 2 have different mean mass metrics (since Experiments 1 and 2 were executed with different nutrient schedules, as in \ref{ssec:grow_procedure}).  To balance the data, only 28 day-old plants were used.  The results are presented in column 4 of Table \ref{tab:anova} and show that our method's metrics are less powerful than the GT masses, but still more powerful than the baseline methods.

\begin{table}
  \centering
  \caption{Plant Harvest Age Distribution} \label{tab:plant_age}
  \vspace*{-1.2em}\# of Samples for Each Age\\[1.2em]
  \scriptsize
  \begin{tabular}{c|cccccccccc}
    Age (days)   & 8 & 15 & 21 & 22 & 23 & 24 & 25 & 26 & 27 & 28 \\
    \hline
    Experiment 1 & 6 & 11 & 11 & ~ & ~ & ~ & ~ & ~ & ~ & 3 \\
    Experiment 2 & ~ & ~ & 3 & 3 & 2 & 3 & 3 & 3 & 3 & 3 \\
  \end{tabular}
\end{table}
\begin{table}
  \caption{ANOVA Statistical Significance Tests}\label{tab:anova}
  \centering
  \begin{tabular}{l|cc|cc}
    \multirow{3}{*}{\centering Metric} & \multicolumn{2}{c|}{p-value for} & \multicolumn{1}{c}{p-value for} \\
      & \multicolumn{2}{c|}{Age Discrimination} & \multicolumn{1}{c}{Nutrient Schedule} \\
      & Exp. 1 & Exp. 2 & Discrimination \\
    \hline
    %
    Fresh Mass (GT)             & 0.\o\o156 & 0.\o\o\o37 & 0.\o\o284 \\
    Dry Mass (GT)               & 0.\o\o137 & 0.\o\o263 & 0.\o\o288 \\
    Surface Area (ours)         & 0.\o\o219 & 0.\o\o352 & 0.\o3134 \\
    Volume (ours)               & 0.\o\o204 & 0.\o\o338 & 0.\o3766 \\
    Baseline 1: Projected Area  & 0.\o\o\o86 & 0.\o2661 & 0.32745 \\
    Baseline 2: Surface Area    & 0.\o\o287 & 0.31166 & 0.32066 \\
    Baseline 2: Volume          & 0.\o\o265 & 0.26535 & 0.28106 \\
  \end{tabular}
\end{table}

\subsection{Point Cloud Occlusion}
We claim that a key advantage of our robot design is that the dexterity of the robot arm allows us to see more of the lettuce plant thereby reducing occlusions (as compared to high-throughput, over-canopy methods).
To assess the degree of occlusion, we used a number of assumptions to generate a metric for ``occlusion proportion''.
Intuitively, we observed that the smallest plants had negligible occlusion so their estimates should be the most accurate.  This is reflected by the slightly convex shapes of the data in Figure \ref{fig:regressions}, indicating that extrapolating the lettuce density from small plants would produce under-estimates for the masses of large plants.
Assuming that underestimates in the mass are due primarily to occlusion, we approximated the occlusion proportion $d\approx 1-\frac{\hat{m}}{m}$ where $m$ is the true mass, $\hat{m}:=\rho X_{est}$ is a na\"ively estimated mass (because it does not compensate for occlusions), and $X_{est}$ is the computed metric from the plant photos (surface area, volume, or projected area).  We also approximated $d$ to be proportional to the depth of a cube with the same mass and density as the lettuce: $d\approx k m_{est}^{1/3}$, where $k$ (``occlusion coefficient'') indicates a method's susceptibility to occlusion (smaller $k$ is better).  Finally, we can derive a new equation for the occlusion-compensated mass estimate:
\begin{align}
  \hat{m}_{occ} &= \frac{\hat{m}}{1-d}
                = \frac{\rho X_{est}}{1- k\left(\rho X_{est}\right)^{1/3}}. \label{eq:occlusion_compensated}
\end{align}
By running a new regression using this model instead of the linear one in \ref{ssec:regression}, we can compare $k$ for different methods, where small $k$ indicates less occlusion. 

\begin{figure}
  \vspace*{-2.5em} 
  \centering
  \includegraphics[width=0.6\linewidth]{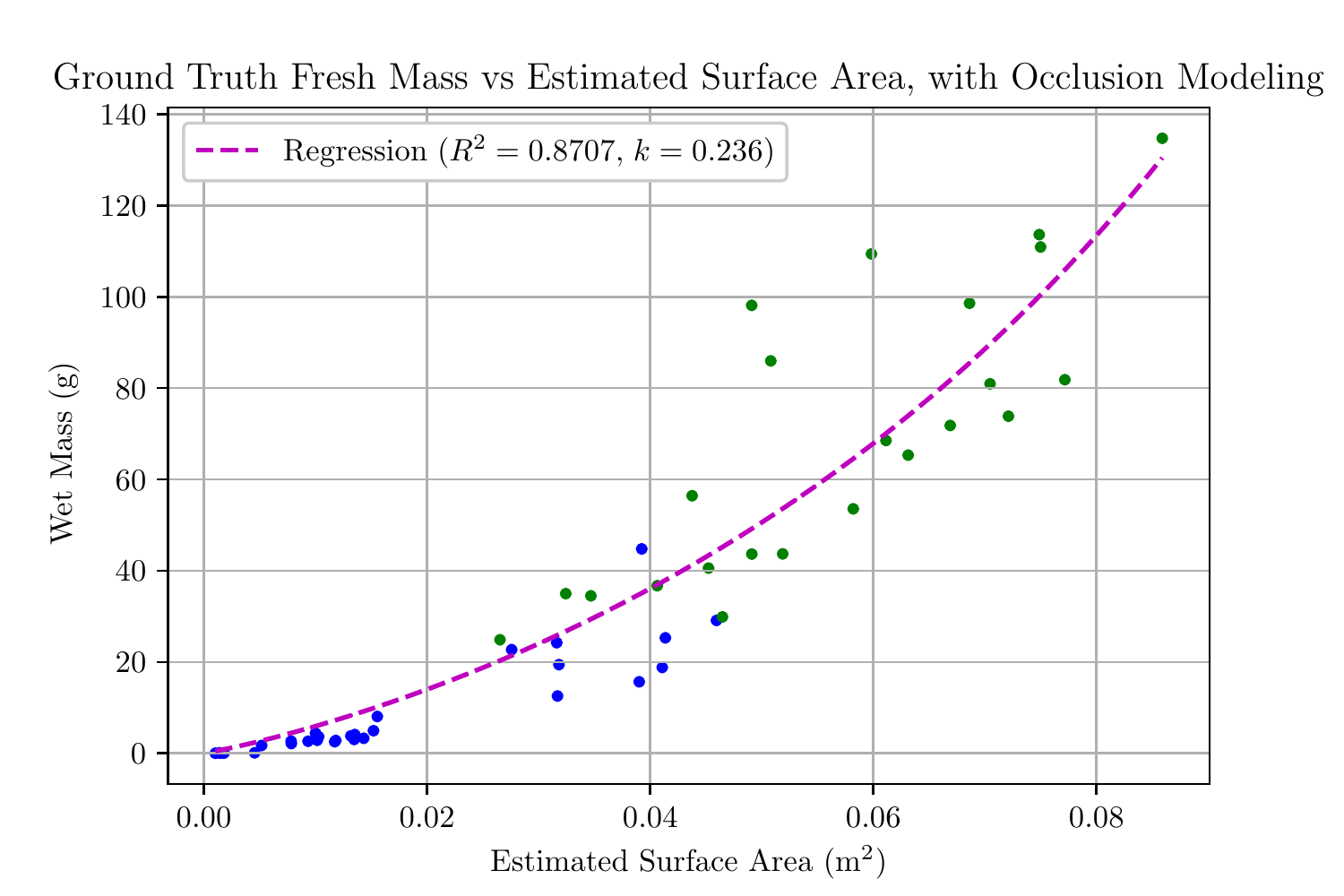}
  \includegraphics[width=0.6\linewidth]{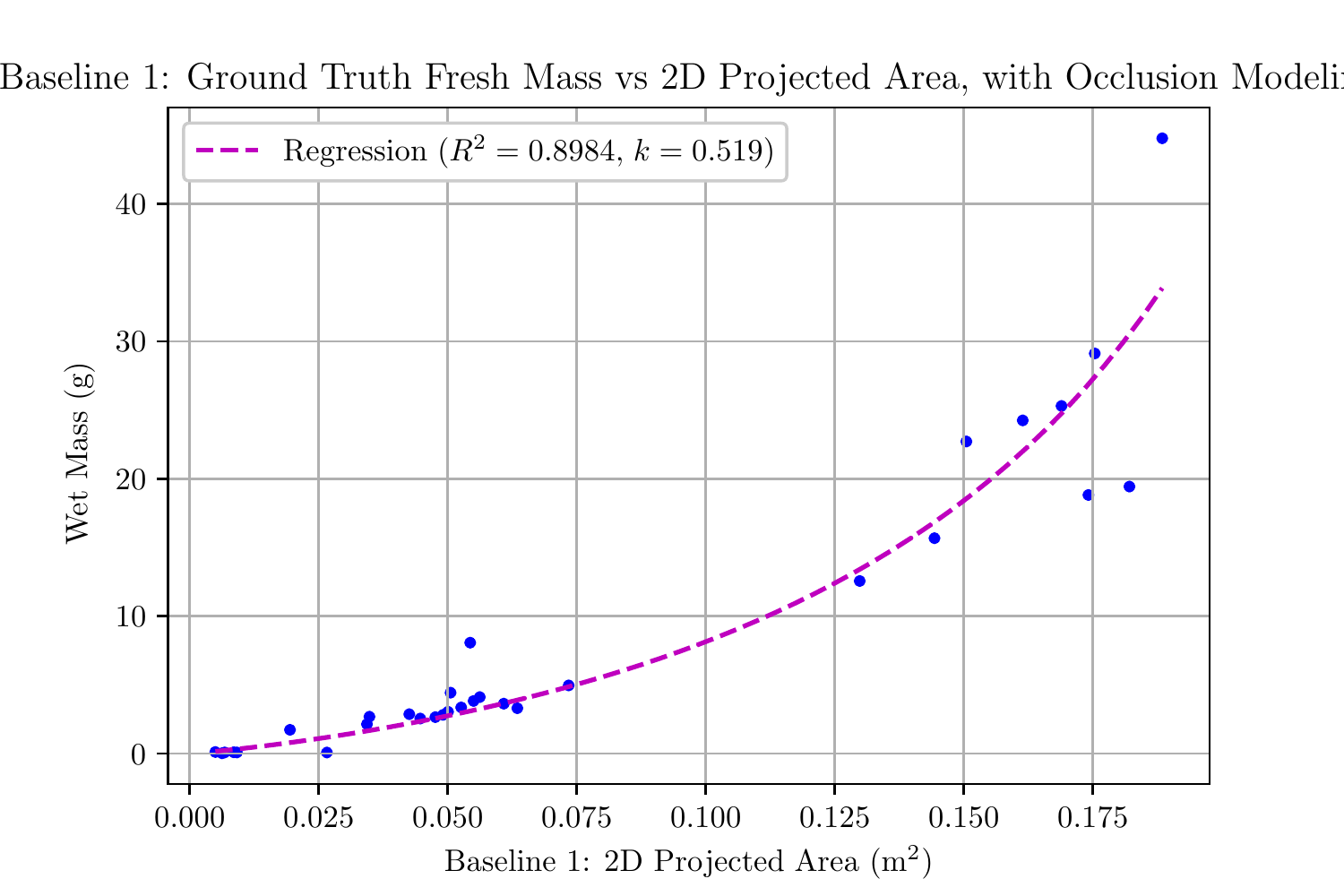}
  \caption{Representative regression results using the occlusion-compensated model (Eq. \eqref{eq:occlusion_compensated}) to estimate fresh mass (ground truth) from (top:) the surface area computed from SfM and (bottom:) projected area estimate from a single top-down photo of each plant, suggest that our method is less susceptible to occlusion.  Lower $k$ signifies less occlusion (better).}
  \label{fig:regressions_occ}
\end{figure}

The occlusion-compensated regression results show that our method has the smallest occlusion coefficient, indicating that the dexterity of the robot arm was helpful in reducing occlusions and capturing a more complete reconstruction of the plant.
Representative regressions are shown in Figure \ref{fig:regressions_occ} and the occlusion coefficients are shown in Table \ref{tab:occlusion}.

\begin{table}
  \centering
  \caption{Susceptibility of Different Methods to Occlusion}\label{tab:occlusion}
  \begin{tabular}{l|cc}
    \multirow{3}{*}{Estimation Method} & \multicolumn{2}{c}{Occlusion coefficient, $k$ (\SI{}{\per\g})} \\
                                       & \multicolumn{2}{c}{(lower is better)}\\
                                       & GT: Fresh Mass & GT: Dry Mass \\
    \hline
    Surface Area & \bf 0.236 & \bf 0.593 \\
    Volume & 0.261 & 0.659 \\
    Baseline 1: Projected Area & 0.519 & 0.883 \\
    Baseline 2: Surface Area & 0.333 & 0.680 \\
    Baseline 2: Volume & 0.350 & 0.743 \\
  \end{tabular}
\end{table}

\subsection{Point Cloud Visualizations and Qualitative Descriptions}
\begin{figure}
  \centering
  \includegraphics[width=0.49\linewidth,trim={0 450px 0 0}, clip]{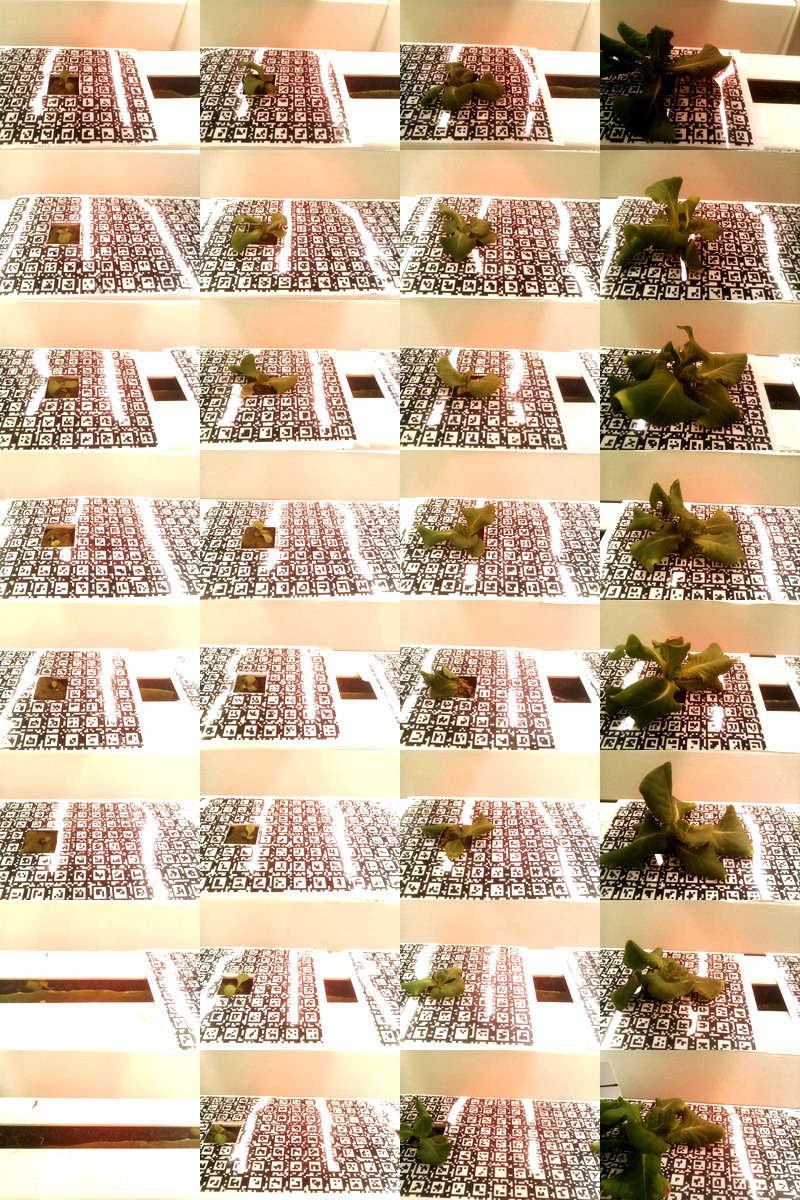}
  \includegraphics[width=0.49\linewidth,angle=180,origin=c]{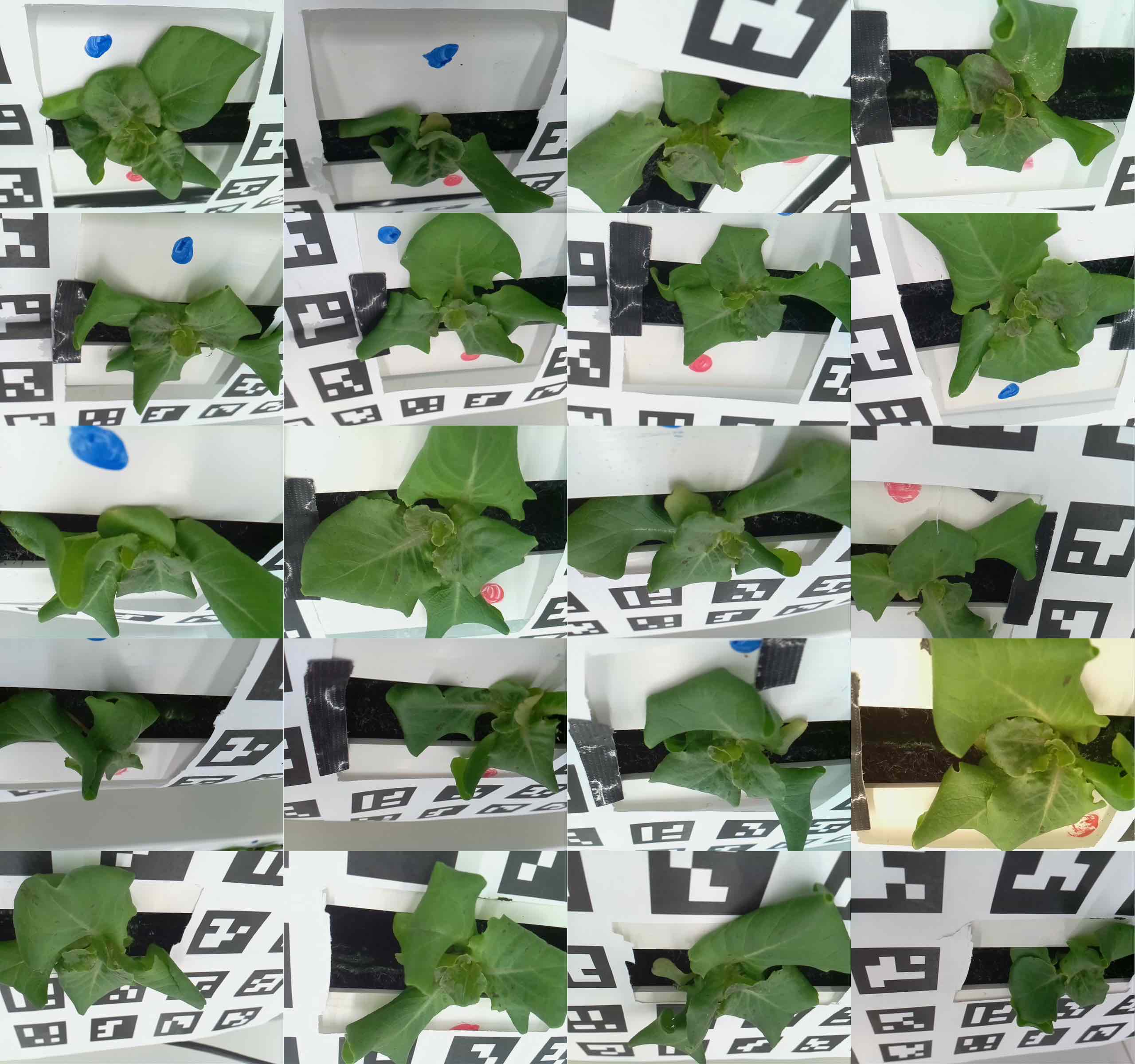}
  \caption{Left: Example photos from our plant dataset depict the consistency with which our robot is able to photograph different plants from the same relative camera angle.  Right: In contrast, photos taken using only the robot arm (Baseline 3) are inconsistent and labor-intensive.}
  \label{fig:plants_mosaic}
\end{figure}

\begin{figure}
  \centering
  \includegraphics[width=0.49\linewidth, trim={300 300 300 300}, clip]{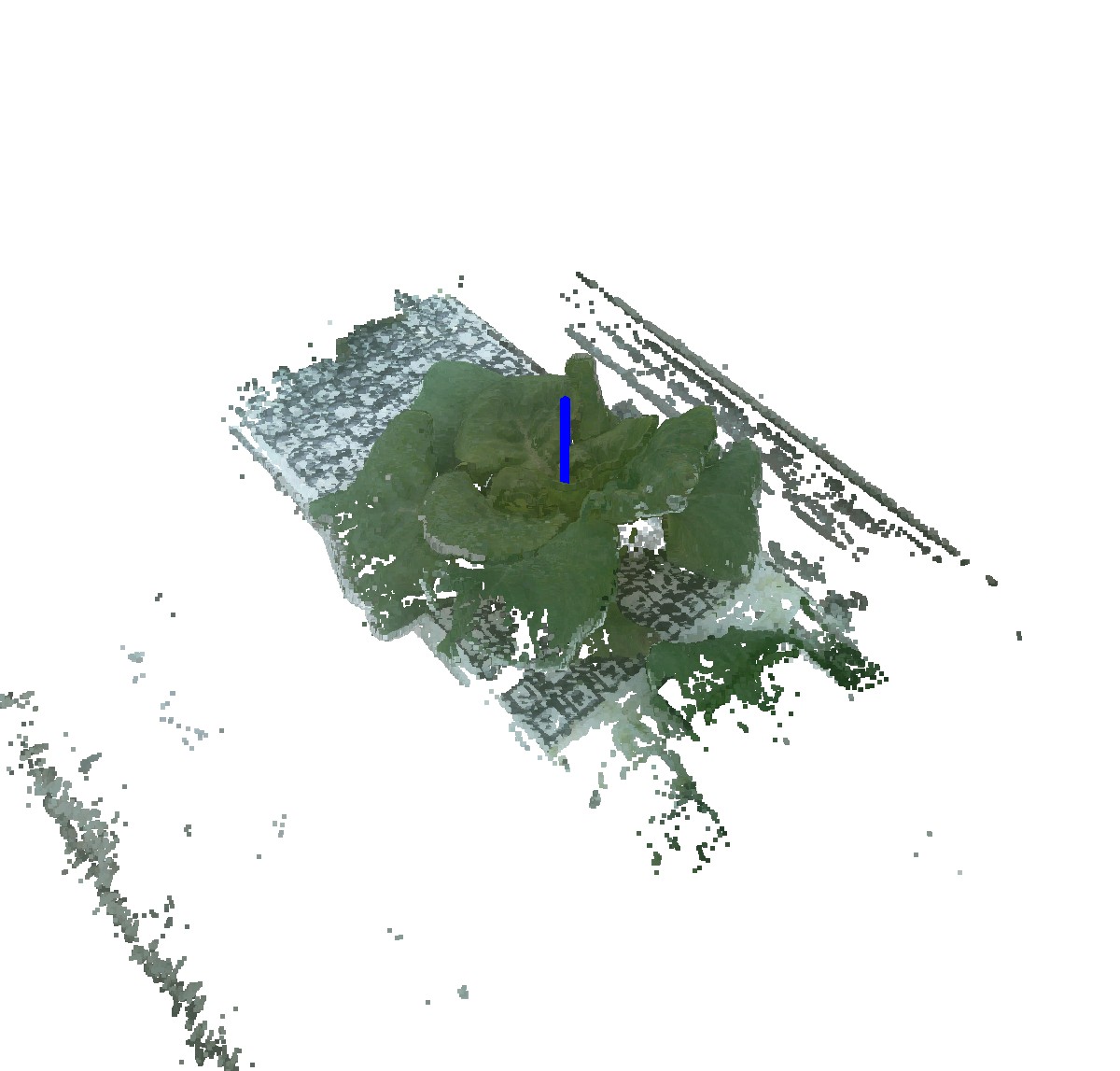}
  \includegraphics[width=0.49\linewidth, trim={320 320 313 320}, clip]{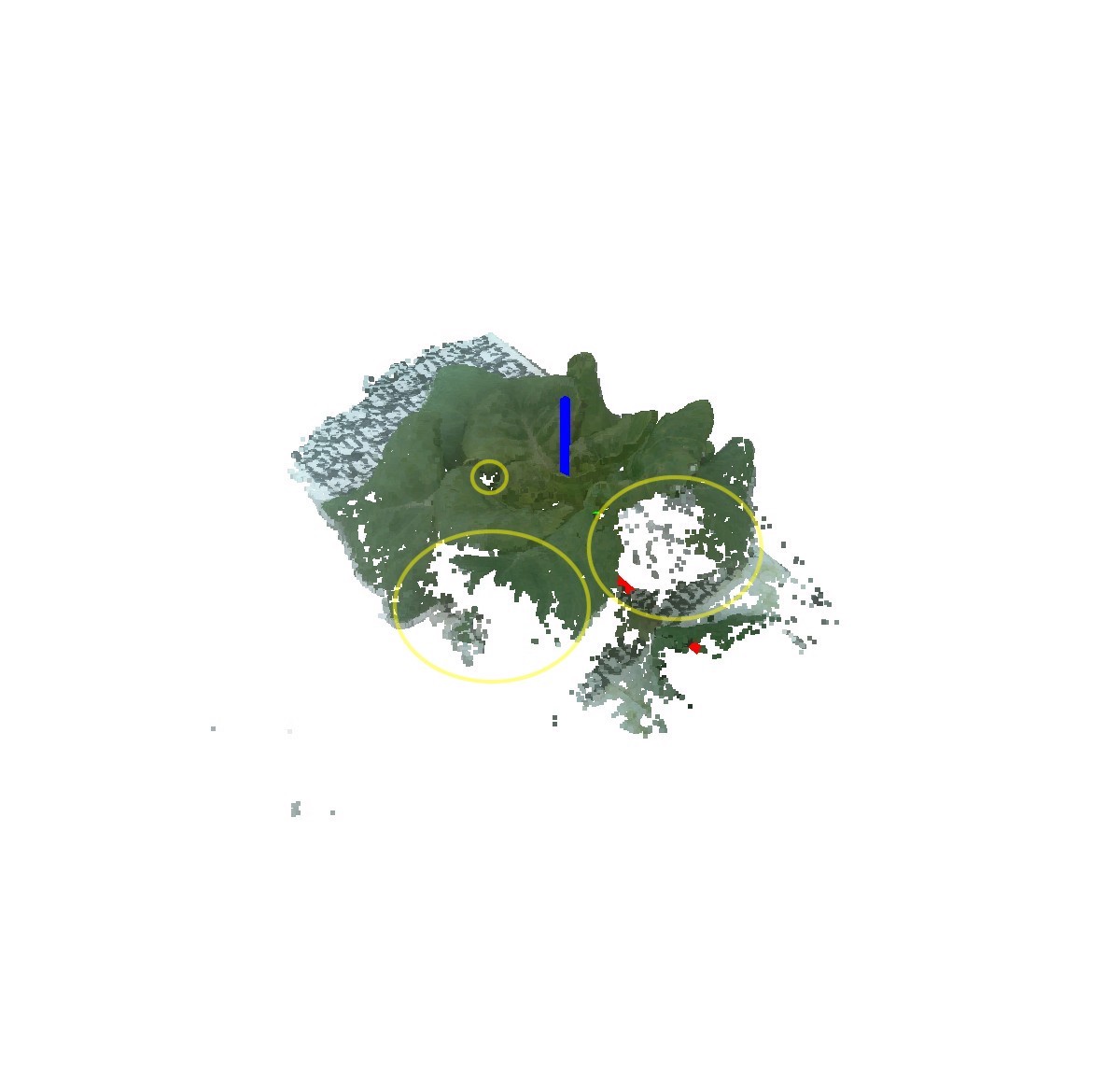}
  \caption{Dense reconstructions of an example lettuce plant using all the photos (left) vs a subset of photos according to Baseline 2 (right) collected by our robot show that the additional camera viewpoints enabled by our hybrid CDPR+arm design are helpful in reducing occlusions (circled in yellow) and capturing a more complete reconstruction of the plant.} \label{fig:pointclouds}
\end{figure}

Both the photos (Fig. \ref{fig:plants_mosaic}) and resulting point clouds (Fig. \ref{fig:pointclouds}) show that our robot is capable of capturing lettuce plant photos from a variety of viewpoints for use in non-destructively estimating plant mass.
Fig. \ref{fig:pointclouds} identifies gaps in the reconstruction due to occlusion that would result from an over-canopy approach (Baseline 2), thereby supporting the claim that the additional dexterity afforded by our hybrid CDPR with robot arm is helpful in reducing occlusions.

A comparison between photos taken by our robot design and photos taken by Baseline 3 (the arm alone), depicted in Fig. \ref{fig:plants_mosaic}, demonstrates the improved consistency with which our robot can position the camera for photos.
Camera pose relative to the plant center is highly variable due to human placement error, and the human labor/oversight required is significant (6-9 plants / hour vs no supervision required for the CDPR).
3D reconstructions for Baseline 3 are not shown: aside from the occasional missing points due to improperly framed photos (human error), the reconstructions were similar to those obtained using our robot design.


\section{CONCLUSIONS and FUTURE WORK}
In this work, we designed and validated a medium-throughput, high-accuracy plant monitoring robot.
We demonstrated its utility by estimating the masses of leafy plants across 2 growth cycles and comparing against 2 baselines simulating higher-throughput systems with fewer available camera viewpoints.  We achieved comparable statistical power to ground-truth mass measurements, signifying scientific utility.  We also achieved consistently superior performance to the baselines, signifying that, compared to high-throughput approaches to plant monitoring, we achieve higher accuracy in return for our slightly reduced throughput.
As compared to prior approaches prioritizing (1) high-throughput but limited viewpoints or (2) low-throughput but many viewpoints, our approach strikes a balance to autonomously collect large numbers of plant photos from a diverse, repeatable set of viewpoints.


Future works include applying visual servo-ing and other real-time computer vision processing techniques to collect better framed photos, analyzing the minimal mechanical design required of the robot arm to achieve high accuracy, and applying more modern techniques for estimating plant properties using computer vision.



\addtolength{\textheight}{-2cm} 
\section*{ACKNOWLEDGMENTS}
We thank Andrew Sharkey, Thomas Igou, and Teagan Groh for their growing assistance and horticultural expertise.


\IEEEtriggeratref{36}
\bibliographystyle{IEEEtran}
\bibliography{references/plant_3D_reconstruction_ris, references/custom, references/references, references/gerry.bib}

\begin{thebibliography}{10}
\bibitem{Sharkey23_pilot_site_tbd}
A.~Sharkey and Y.~Chen, ``To be published: analysis on the effect of potassium
  starvation on lettuce growth.'' This citation will be filled in before
  camera-ready.
\end{thebibliography}

\end{document}